\documentclass[runningheads]{llncs}

\usepackage[mobile]{eccv}

\usepackage{eccvabbrv}

\usepackage{graphicx}
\usepackage{booktabs}
\usepackage[table]{xcolor} 
\usepackage{booktabs}
\usepackage{graphicx}
\usepackage[accsupp]{axessibility}  

\usepackage[pagebackref,breaklinks,colorlinks,citecolor=eccvblue]{hyperref}

\usepackage{orcidlink}
\definecolor{darkgreen}{rgb}{0.0, 0.42, 0.24}

\definecolor{darkred}{RGB}{128, 0, 0}
\definecolor{darkblue}{RGB}{0, 0, 128}

\begin{document}

\title{AnyRecon: Arbitrary-View 3D Reconstruction with Video Diffusion Model} 


\author{Yutian Chen\inst{1,2} \and
Shi Guo\inst{1,\dagger} \and
Renbiao Jin\inst{1,3} \and
Tianshuo Yang \inst{1,4} \and
Xin Cai \inst{1,2} \and
Yawen Luo \inst{2} \and
Mingxin Yang\inst{1,3} \and
Mulin Yu \inst{1} \and
Linning Xu \inst{1,2} \and
Tianfan Xue \inst{2,1,5} }
\renewcommand{\thefootnote}{\fnsymbol{footnote}}
\footnotetext[0]{\hspace{-1.1em} \textsuperscript{\dag} indicates corresponding author.}

\authorrunning{Y. Chen et al.}

\institute{ }

\maketitle

\begin{center}
{\small 
\textsuperscript{1}Shanghai AI Lab \quad 
\textsuperscript{2}CUHK MMLab \quad 
\textsuperscript{3}Shanghai Jiao Tong University \\[0.3em]
\textsuperscript{4}The University of Hong Kong \quad 
\textsuperscript{5}CPII under InnoHK
}

\smallskip
{\href{https://yutian10.github.io/AnyRecon/}{https://yutian10.github.io/AnyRecon/}}
\end{center}
\begin{abstract}
  Sparse-view 3D reconstruction is essential for modeling scenes from casual captures, but remain challenging for non-generative reconstruction. Existing diffusion-based approaches mitigates this issues by synthesizing novel views, but they often condition on only one or two capture frames, which restricts geometric consistency and limits scalability to large or diverse scenes. We propose AnyRecon, a scalable framework for reconstruction from arbitrary and unordered sparse inputs that preserves explicit geometric control while supporting flexible conditioning cardinality. To support long-range conditioning, our method constructs a persistent global scene memory via a prepended capture view cache, and removes temporal compression to maintain frame-level correspondence under large viewpoint changes. Beyond better generative model, we also find that the interplay between generation and reconstruction is crucial for large-scale 3D scenes. Thus, we introduce a geometry-aware conditioning strategy that couples generation and reconstruction through an explicit 3D geometric memory and geometry-driven capture-view retrieval. To ensure efficiency, we combine 4-step diffusion distillation with context-window sparse attention to reduce quadratic complexity. Extensive experiments demonstrate robust and scalable reconstruction across irregular inputs, large viewpoint gaps, and long trajectories. 
\end{abstract}

\section{Introduction}
\label{sec:intro}

\begin{figure}[!t]
   \centering
   \includegraphics[width=1\textwidth]{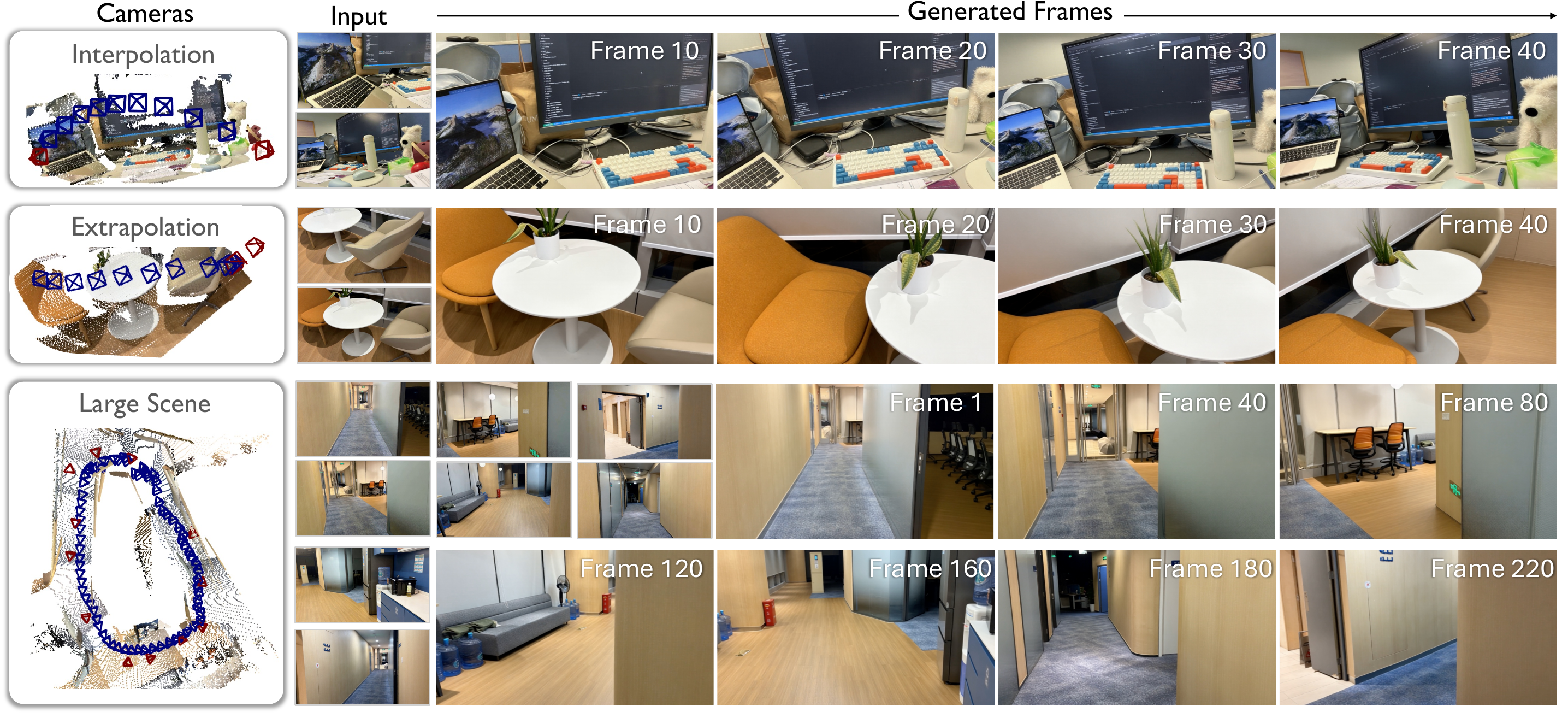}
   \caption{
    {AnyRecon demonstrates robust performance across multiple reconstruction settings: (Top) Interpolation, filling in gaps between distant captured views; (Middle) Extrapolation, synthesizing novel content beyond the observed range; and (Bottom) Large Scene Reconstruction, maintaining consistency across long-trajectory sequences. The camera visualization (left) illustrates the sparse \textbf{\textcolor{darkred}{input poses (red)}} and the dense \textbf{\textcolor{darkblue}{generated path (blue)}}}.
    }
   \label{fig:teaser}
   \vspace{-5mm}
\end{figure}
Novel view synthesis and 3D reconstruction are fundamental problems in computer vision and graphics, enabling applications ranging from immersive virtual environments to augmented reality and visual effects. Recent advances in neural reconstruction methods, including implicit representations such as NeRF~\cite{mildenhall2021nerf} and explicit point-based approaches like 3D Gaussian Splatting~\cite{kerbl20233d}, have demonstrated remarkable visual fidelity.
However, these methods rely on densely sampled multi-view images captured under controlled acquisition setups. While real-world visual data—such as handheld captures or Internet videos—are typically sparse and irregular. Enabling reconstruction from such arbitrary view would allow scalable conversion of everyday captures into explorable 3D scenes.

Recent work try to mitigate this sparse view challenge by using more views created by diffusion-based novel view synthesis. While early efforts employ image generative models to infer 3D structure~\cite{wu2024reconfusion,wu2025difix3d+}, more recent works leverage video generation models to synthesize novel views, as they can better capture cross-view coherence through temporal modeling. One line of work~\cite{yu2024viewcrafter,ren2025gen3c,cao2025uni3c} conditions diffusion models primarily on projected point cloud renderings, providing only one or two captured RGB frames (e.g., the first and last views). Although renderings offer coarse geometric guidance, limited real-image conditioning weakens appearance fidelity and global context, making generation sensitive to incomplete or low-quality geometry. Another line of work~\cite{bai2025recammaster,gao2024cat3d} relies solely on RGB images and camera poses without explicitly incorporating reconstructed geometry into the generation process. By learning geometric consistency implicitly, these methods struggle to maintain precise pose alignment and spatial consistency, limiting their deployment for real-world 3D reconstruction under irregular observations.

In this work, we aim to enable high-quality and large scale 3D reconstruction from sparse inputs. Unlike prior sparse-view diffusion models restricted to only one or two reference views, our diffusion model flexibly conditions on an arbitrary number of captured RGB images alongside point cloud renderings. However, supporting flexible input cardinality introduces many challenges. First, input images may be arbitrary captured with large viewpoint gaps, while existing video diffusion frameworks are suboptimal for non-sequential inputs, as they rely on temporally causal latent compression~\cite{liu2025dreamontage}. Moreover, to reconstruct a large complex scene, it is impossible to fit all input into diffusion model at once. Therefore, a robust iterative reconstruction strategy is required to reconstruct a scene by small segment. These challenges demand a reconstruction framework that preserves fine-grained control while remaining computationally efficient.

To address these challenges, we propose AnyRecon, a scalable framework for sparse-view 3D reconstruction. First, we develop a diffusion-based novel view synthesis model that supports arbitrary and unordered sparse inputs while maintaining explicit geometric control. Specifically, we construct a 3D point cloud from the sparse captures and render it into target viewpoints to serve as geometric conditions. To handle flexible inputs, we maintain a global memory cache by prepending these original captures to the rendered priors in the sequence, thereby enabling long-range conditioning across arbitrary viewpoints. Furthermore, to ensure frame-level correspondence under large viewpoint changes, we remove temporal compression in the latent encoder. At last, to reduce computation for large-scale reconstruction, we adopt a 4-step diffusion distillation strategy and introduce a context-window sparse attention mechanism that restricts attention to local temporal windows and geometry-aligned retrieved views.

Furthermore, to support segment-by-segment reconstruction of a large-scale scene, we introduce a geometry-aware conditioning strategy that couples generation and reconstruction. Specifically, this strategy creates an iterative loop where generated outputs continuously update a shared 3D geometry, which in turn guides the conditioning of subsequent segments. First, we build an explicit 3D Geometry Memory by back-projecting newly generated images into the initial point cloud, enabling incrementally updated geometric memory across trajectory segments. Second, when conditioning a new trajectory segment, we perform geometry-driven view selection from a captured view bank based on geometric contribution and spatial overlap with the current reconstruction, rather than relying on image-level similarity or field-of-view (FOV) heuristics~\cite{yu2025context,li2025vmem,chen2026context}. Together, these components form a closed geometric loop between reconstruction and generation, ensuring that diffusion is guided by spatially informative observations and improving robustness under occlusion and complex scene layouts.

Extensive benchmarks demonstrate that AnyRecon delivers superior results compared to state-of-the-art baselines. As shown in Fig.~\ref{fig:teaser}, our method facilitates seamless view interpolation and long-range extrapolation across diverse, large-scale scenes (over 200 frames), maintaining high fidelity despite sparse and irregular input captures.

Our key contributions are summarized as follows:
\begin{itemize}
    \item \textbf{A flexible sparse-view reconstruction framework}. We propose a video-diffusion-based approach that supports arbitrary and unordered conditioning views, enabling robust 3D reconstruction.
    \item \textbf{A geometry-aware conditioning design.} We couple generation and reconstruction via a 3D Geometry Memory with back-projection and geometry-driven capture-view retrieval, ensuring spatially grounded diffusion guidance.
    \item \textbf{An efficient diffusion architecture.} By removing temporal compression and adopting diffusion distillation with block sparse attention, our method generalizes across varying numbers of input views while maintaining computational efficiency.
\end{itemize}

\section{Related Work}
\label{sec:related}

\subsection{Traditional Sparse-View Reconstruction.} 
Sparse-view reconstruction is inherently ill-posed due to the vast unobserved regions and geometric ambiguities arising from limited inputs. 
Early efforts addressed this by incorporating various geometric priors and regularization techniques. 
For instance, FreeNeRF~\cite{yang2023freenerf} and RegNeRF~\cite{niemeyer2022regnerf} employ frequency-domain regularization and depth smoothness constraints to stabilize optimization when only sparse views are available. 
Other approaches focus on leveraging auxiliary supervision from pre-trained models to provide additional scene constraints. 
Specifically, SPARF~\cite{sparf2023} utilizes correspondence field and optical flow to enforce multi-view consistency, while MonoSDF~\cite{yu2022monosdf} and DS-NeRF~\cite{kangle2021dsnerf} integrate monocular depth and normal maps as supplementary signals to refine surface geometry. 

While these methods improve reconstruction to some extent, they rely on the limited information present in the sparse inputs and often struggle to synthesize plausible details in large disoccluded areas. 
This motivates the shift toward the diffusion-based approaches discussed above, which leverage large-scale generative priors to hallucinate consistent geometry and appearance beyond the captured observations.

\subsection{3D Reconstruction with Diffusion Model}
Recent advancements in diffusion models have significantly propelled 3D and 4D generation. 
Pioneer works such as ReconFusion~\cite{wu2024reconfusion} and FreeNeRF~\cite{yang2023freenerf} supervise novel views by sampling from diffusion priors during the reconstruction process. 
Specifically, ReconFusion~\cite{wu2024reconfusion} utilizes a diffusion model conditioned on sparse inputs to predict pseudo-ground-truth images for novel views, which are then used to optimize NeRF or 3DGS. 
To improve optimization efficiency, Deceptive-NeRF~\cite{liu2023deceptive} and 3D-GS Enhancer~\cite{gao2024cat3d} first render coarse pseudo-images from a sparse-view-reconstructed representation and refine these views using diffusion models, avoiding the need to query the diffusion prior at every step. 
However, these image-based approaches often lack cross-view geometric coherence and necessitate computationally expensive iterative refinements.

To address the temporal and spatial consistency issues of image-based models, recent research has shifted towards video-based generators. 
One category of methods~\cite{bai2025recammaster,gao2024cat3d} relies primarily on RGB images and camera poses to implicitly learn spatial consistency. While flexible, these methods struggle with precise pose alignment without explicit geometric guidance. 
Alternatively, geometry-aware video generators~\cite{yu2024viewcrafter,ren2025gen3c,cao2025uni3c,chen2025scenecompleter} incorporate projected point cloud renderings to provide coarse structural priors. These point clouds are typically estimated using pretrained geometry reconstruction models~\cite{wang2025pi,wang2025vggt,wang2024dust3r,wang2025continuous}, which infer sparse or dense 3D structures from input images before projection into the diffusion pipeline.
Nevertheless, these diffusion model typically condition on a very limited number of captured frames (e.g., only the first and last), which limits their ability to capture global scene context and fine-grained appearance fidelity. 
In contrast, our AnyRecon utilizes a global scene memory to incorporate an arbitrary number of reference views $\mathcal{I}_{sel}$ and enforces strict spatial alignment through channel-wise concatenation of visibility masks $M_t$ and rendered observations $I_{render}$, closing the gap between generative priors and explicit 3D geometry.

\section{Method}
\label{sec:method}

\begin{figure}[!t]
   \centering
   \includegraphics[width=1\textwidth]{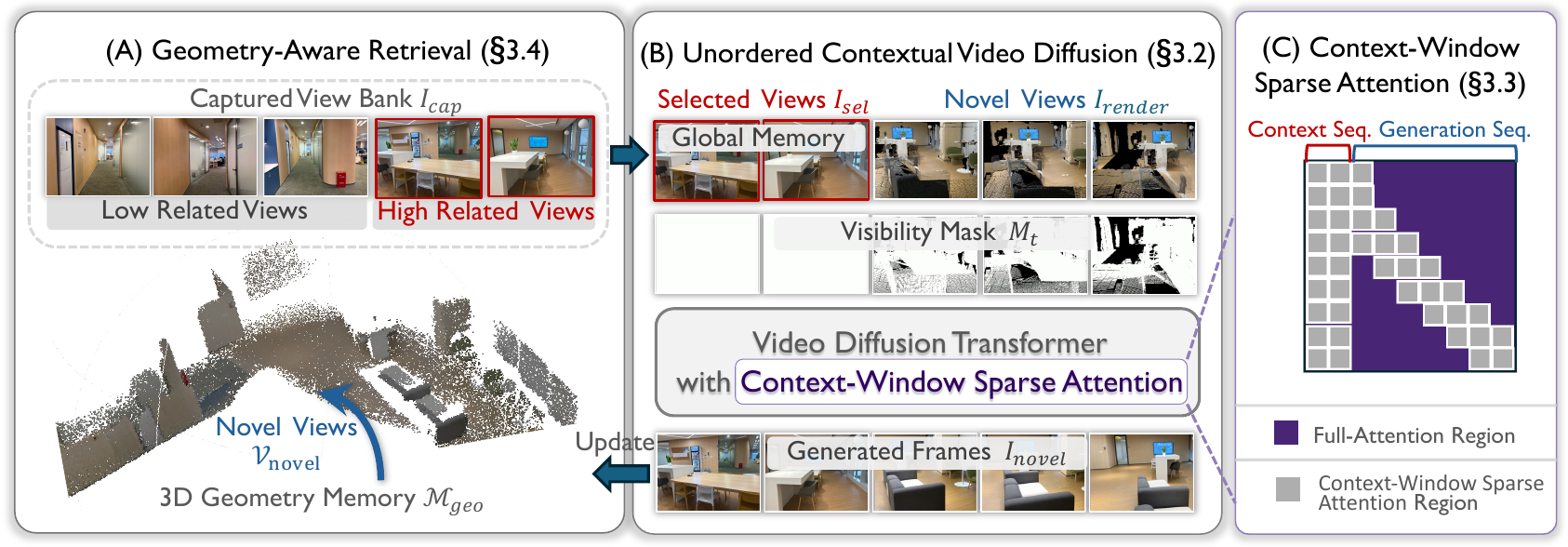}
   \caption{
    \textbf{Pipeline of AnyRecon.}
    Given arbitrary sparse input views organized in a capture view bank $\mathcal{I}_{cap}$, we perform geometry-aware retrieval to select spatially informative views for each novel trajectory segment. 
    The selected views $\mathcal{I}_{sel}$, together with geometry renderings under target viewpoints $\mathcal{I}_{render}$, are fed into a video diffusion transformer equipped with context-window sparse attention for scalable long-range conditioning. 
    The generated novel views are then used to update the 3D geometry memory $\mathcal{M}_{geo}$, forming a closed loop between generation and reconstruction.
    }
   \label{fig:pipeline}
   \vspace{-5mm}
\end{figure}

\subsection{Overview}
\label{sec:overview}
We present \textbf{AnyRecon}, a framework for sparse-view 3D reconstruction that supports arbitrary and unordered inputs while preserving geometric consistency across long viewpoint changes. Our method alternates between (1) diffusion-based trajectory generation to create more views and (2) geometry refinement using the generated views. This forms a closed loop that progressively reconstructs the scene, enabling scalable processing of long trajectories and large-scale inputs.

As illustrated in Fig.~\ref{fig:pipeline}, our framework operates in three key stages to form the generation-reconstruction loop. 
First, \textit{Initial Geometry Construction:} All input views are organized into a captured view bank $\mathcal{I}_{cap}$, from which an initial 3D geometry memory $\mathcal{M}_{geo}$ is established via a feed-forward point map estimation method (e.g., VGGT~\cite{wang2025vggt} or $\pi^3$~\cite{wang2025pi}). 
Second, \textit{Novel View Generation:} To synthesize novel views on user-specified trajectory $V_{novel}$, we chop the entire trajectory into small for efficiency. For each segment, we perform geometry-aware retrieval (\S\ref{sec:geometry}) to select views $\mathcal{I}_{sel}$ important for reconstructing this segment from all capture views $\mathcal{I}_{cap}$. These selected views, along with point-cloud renderings $I_{render}$ and visibility masks $M_t$ derived from the current $\mathcal{M}_{geo}$, are fed into our proposed unordered contextual video diffusion (\S\ref{sec:diffusion}) to synthesize novel views $\hat{I}_{novel}$ on the trajectory. To mitigate computational complexity, we incorporate context-window sparse attention and an efficient 4-step sampling strategy (\S\ref{sec:sparse}). 
Third, \textit{Geometry Updating:} The geometry reconstructed from the newly synthesized views $\hat{I}_{novel}$ is extracted to update the global memory $\mathcal{M}_{geo}$ (\S\ref{sec:geometry}). This updated geometry is subsequently fed back into the next segment's retrieval and generation steps, completing the iterative reconstruction loop.

\subsection{Unordered Contextual Video Diffusion}
\label{sec:diffusion}
To achieve robust reconstruction from diverse inputs, AnyRecon transitions from a standard sequential video generator to a geometry-conditioned diffusion model. 
Specifically, this module takes the retrieved reference views $\mathcal{I}_{sel}$ and the rendered geometric guidance $I_{render}$ under target viewpoints $V_{novel}$ as joint contextual inputs to synthesize a sequence of high-fidelity novel views $\hat{I}_{novel}$. 
To ensure both precise spatial alignment and awareness of occlusions, the target noisy latents are {concatenated along the channel dimension} with the rendered point-cloud observations $I_{render}$ and their corresponding visibility masks $M_t$, both of which are derived from the 3D geometry memory $\mathcal{M}_{geo}$. Beyond this spatial geometry conditioning, effectively utilizing sparse and unordered inputs requires breaking the strict temporal continuity assumptions inherent in standard video diffusion models. To fully decouple the generation process from temporal dependencies and handle arbitrary viewpoint gaps, we introduce two key architectural innovations: a global scene memory for flexible context injection, and a non-compressive latent encoding to prevent spatial-temporal feature entanglement.

\textbf{Global Scene Memory.} To support an arbitrary number of conditioning views without being constrained by fixed-length input buffers, we introduce a \textit{Global Scene Memory} mechanism as shown in Fig. \ref{fig:pipeline}(B). 
{Specifically, the retrieved reference views $\mathcal{I}_{sel}$ are set in the beginning of each chunk and serve as a persistent global key–value (KV) memory cache within the video diffusion transformer.} (See Sec. \ref{sec:ablation} for further comparative analysis.)

{Instead of modeling captured and target views as temporally adjacent frames in a single sequence~\cite{yu2024viewcrafter,cao2025uni3c}, this design treats conditioning capture views as a flexible and queryable 3D memory, enabling generation along arbitrary spatial trajectories independent of the capture sequence.}
The model then generates novel views based on this memory, enabling spatially consistent reconstruction.

\begin{figure}[!t]
   \centering
   \includegraphics[width=1\textwidth]{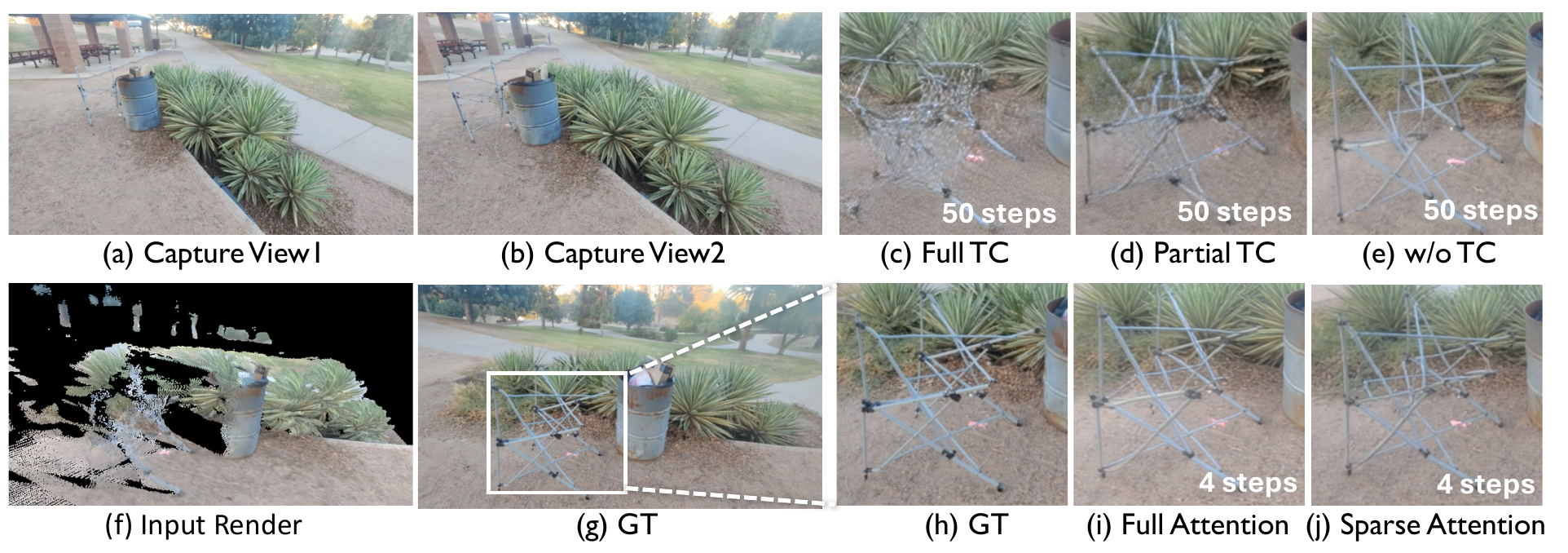}
   \caption{\textbf{Ablation on temporal compression (TC), 4-step distillation and sparse attention.} Full temporal compression follows Wan by keeping only the first frame uncompressed while compressing subsequent frames (e.g., ×4), whereas partial temporal compression compresses only the rendered maps and keeps the captured input views uncompressed to preserve accurate geometric cues. }
   \label{fig:ablation tc}
   \vspace{-5mm}
\end{figure}

\textbf{Non-Compressive Latent Encoding.} Traditional video diffusion models~\cite{wan2025} often use temporal compression (e.g., 3D-VAEs) to reduce dimensionality, which relies on an assumption of temporal smoothness. However, this prior fails in sparse-view scenarios where large viewpoint gaps break the continuity between adjacent frames. Compressing across time in such irregular sequences causes feature entanglement between disparate views, obscuring the precise spatial-temporal alignment necessary for reconstruction as shown in Fig.~\ref{fig:ablation tc} (c) or (d). 

To overcome this limitation, AnyRecon employs Non-Compressive Latent Encoding. By using a frame-wise 2D VAE, we bypass temporal pooling and preserve the one-to-one mapping between latent tokens and pixel coordinates, enabling robust geometry-aware synthesis even with wide-baseline inputs.

\subsection{Efficient Sparse Attention and 4-Step Diffusion Sampling}
\label{sec:sparse}
To maintain high-fidelity synthesis across extended sequences while ensuring computational efficiency, we introduce two key optimizations: a {context-window sparse Attention} mechanism to handle the expanded token space and a {4-step diffusion sampling} strategy to accelerate the generation process. 

\textbf{Context-Window Sparse Attention.} Although the non-compressive encoding and global scene memory enhance rendering quality, they significantly expand the sequence length $L$, resulting in prohibitive $O(L^2)$ complexity. To mitigate this, we introduce a context-window sparse attention mechanism (Fig.~\ref{fig:pipeline}(C)) where each frame in the target trajectory $I_{novel}$ restricts its receptive field to a local temporal window and a selectively retrieved subset of geometry-aligned reference views $\mathcal{I}_{sel}$. This mechanism focuses the model's capacity on visually relevant regions, ensuring scalability for large-scale scenes.

\textbf{4-Step Diffusion Sampling.}
To accelerate the inference of the Wan video diffusion model, we employ Distribution Matching Distillation ~\cite{yin2024onestep,yin2024improved} to distill the pre-trained model into a student network capable of high-quality generation in just 4 steps. We discretize the continuous noise schedule into a fixed trajectory $T_{steps} = \{1000, 750, 500, 250, 0\}$. The optimization objective minimizes the Kullback-Leibler (KL) divergence between the student's generated distribution and the real distribution, approximated via the score difference between a frozen teacher and a trainable critic. To implement this, the generator loss $\mathcal{L}_{gen}$ is formulated as a pseudo-regression objective with a stop-gradient ($\text{sg}$) operator:
\begin{equation}
    \mathcal{L}_{gen} = \mathbb{E}_{z_t, t} \left[ \frac{1}{2} \left\| \hat{x}_\theta(z_t) - \text{sg} \left( \hat{x}_\theta(z_t) + \eta \frac{\hat{x}_\psi(z_t) - \hat{x}_\phi(z_t)}{\sigma_{\text{norm}}} \right) \right\|^2_2 \right],
\end{equation}
where $\hat{x}_\theta, \hat{x}_\psi,$ and $\hat{x}_\phi$ denote the denoised predictions ($\hat{x}_0$) derived from the student, teacher, and critic respectively; $\eta$ is the step size, and $\sigma_{\text{norm}}$ acts as a time-dependent normalization factor. 
Concurrently, the critic is optimized via a standard denoising score matching objective on the student's generated samples: $\mathcal{L}_{critic} = \mathbb{E}_{z_t, t} [ \| \hat{x}_\phi(z_t) - x_{\text{clean}} \|^2_2 ]$, where $x_{\text{clean}}$ is the original noise-free output produced by the student. 
To stabilize the training dynamics, we employ an alternating update schedule between the student generator and the critic.

{Together, these optimizations achieve up to a 20$\times$ speedup in generation over the vanilla diffusion implementation, without obvious rendering degradation.}



\begin{figure}[!t]
   \centering
   \includegraphics[width=1\textwidth]{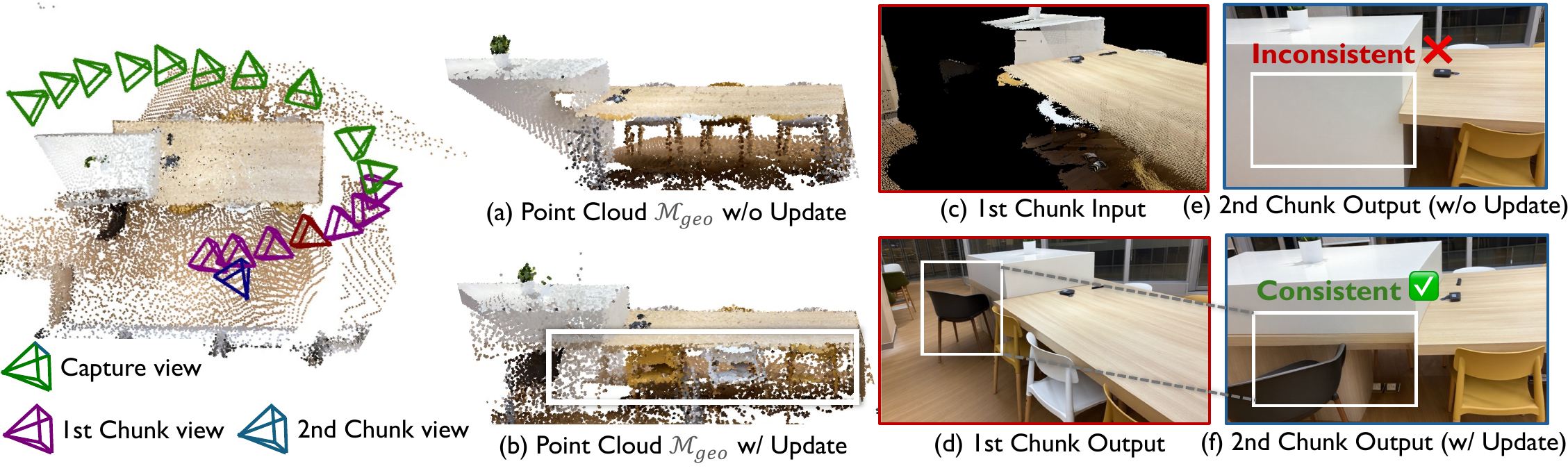}
   \caption{\textbf{Explicit 3D Geometry Memory Update.} Without memory update, newly generated trajectory segments are not integrated into the reconstructed point cloud, leading to incomplete geometry and inconsistent rendering in subsequent chunks. Our explicit memory incrementally integrates generated views into the point cloud, maintaining coherent scene structure across trajectory segments.}
   \label{fig:update}

\end{figure}
\subsection{Geometry-Aware Conditioning Strategy}
\label{sec:geometry}
To support long-trajectory generation and maintain scene-level consistency, we couples the diffusion process with an explicit 3D representation. This coupling forms a recursive loop: newly generated views provide the visual data to expand the 3D reconstruction, while the updated geometry offers precise spatial anchors to guide subsequent generation.

\textbf{3D Geometry Memory Update.} Fig.~\ref{fig:update} illustrates the critical role of our explicit 3D geometry memory $\mathcal{M}_{geo}$. 
Without updating $\mathcal{M}_{geo}$ with newly reconstructed points, newly generated trajectory segments are not integrated into the global scene representation. 
Consequently, conditioning subsequent generation stages on an incomplete $\mathcal{M}_{geo}$ (Fig.~\ref{fig:update}(a)) results in a significant visual and geometric mismatch between previously synthesized views (Fig.~\ref{fig:update}(d)) and those generated in later stages (Fig.~\ref{fig:update}(e)).

To address this, we maintain $\mathcal{M}_{geo}$ as an incrementally updated point cloud that evolves alongside the generation process. 
After synthesizing a segment of the novel trajectory, we employ the feed-forward point map estimation model $\pi^3$~\cite{wang2025pi} to extract 3D geometry from the generated views together with the original ones. 
This newly reconstructed geometry then replaces the existing memory $\mathcal{M}_{geo}$. As shown in Fig.~\ref{fig:update}(b), this fusion successfully recovers missing scene details, such as the chair’s structure.

By iteratively integrating points from generated frames, $\mathcal{M}_{geo}$ evolves into a spatially consistent backbone that anchors each new segment to the global structure. This explicit update mechanism prevents error accumulation, effectively mitigating geometric drift across extended trajectories—as evidenced by the alignment between the early and late stages shown in Fig.~\ref{fig:update}(d) and (f).

\begin{figure}[!t]
   \centering
   \includegraphics[width=1\textwidth]{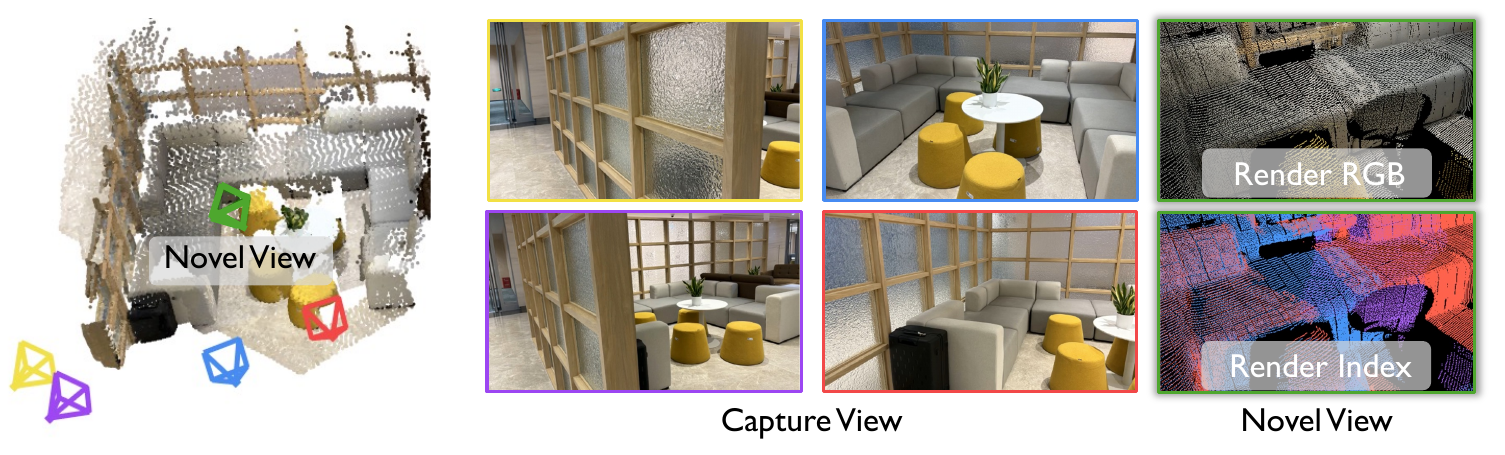}
   \caption{\textbf{Geometry-aware Memory Retrieval.} While FOV- or similarity-based methods would select all four views, our geometry-aware retrieval accounts for 3D spatial overlap and visibility. {In the render index map (right), each color corresponds to a source view, representing its geometric contribution to the target perspective.} This mechanism effectively excludes occluded views (e.g., the yellow view) that provide no valid support, leading to more reliable conditioning for generation.}
   \label{fig:choose}

\end{figure}

\textbf{Geometry-Driven View Selection.} 
When reconstructing scene-level environments, the captured view bank $\mathcal{I}_{cap}$  often contains a massive number of images, making it infeasible to input all reference views into the video diffusion model simultaneously. Therefore, selecting an informative subset is critical for synthesis fidelity and computational efficiency. Specifically, incorporating spatially irrelevant reference views introduces redundant conditioning that distracts the model's spatial reasoning and increases inference latency.
Conversely, omitting highly relevant reference views leads to under-constrained generation, causing the synthesized sequence to deviate from the ground-truth observations.

Fig.~\ref{fig:choose} illustrates our retrieval strategy. 
Given a target novel viewpoint (left), conventional FOV- or similarity-based retrieval would select all capture views due to apparent angular or appearance proximity. 
However, such heuristics ignore occlusion and true geometric support. Instead, we perform {geometry-driven retrieval} based on the current 3D geometry memory $\mathcal{M}_{geo}$. 
Specifically, we render $\mathcal{M}_{geo}$ from the target viewpoint to generate a visibility index map (Fig.~\ref{fig:choose}, right), which identifies the source-view attribution for every visible 3D point.

This visibility map allows us to quantify the geometric contribution of each capture view to the current target perspective. Formally, let $\mathcal{C} = \{(I_i, P_i)\}$ denote the set of capture views and their poses. 
For each candidate view $i$, we compute how many of the visible points under the target viewpoint originate from view $i$ during reconstruction as:

\begin{equation}
    s_i = \frac{|\mathcal{V}_{novel} \cap \mathcal{S}_i|}{|\mathcal{V}_{novel}|},
\end{equation}
where $\mathcal{V}_{novel}$ denotes the set of geometry points visible from the target viewpoint, and $\mathcal{S}_i$ represents the subset of points in $\mathcal{M}_{geo}$ reconstructed from capture view $i$. Views that contribute few or no visible points (e.g., the occluded yellow view in Fig.~\ref{fig:choose}) receive low scores and are filtered out. 
We select the top-$k$ views according to $\{s_i\}$ as conditioning inputs for the diffusion model. By conditioning retrieval on target-view visibility rather than appearance similarity, our method ensures that selected views provide direct geometric support for the current generation. 
This visibility-aware mechanism improves robustness under occlusion and complex spatial layouts, leading to more reliable novel view synthesis.

\section{Experiments}
\label{sec:experiments}

\begin{figure}[!t]
   \centering
   \includegraphics[width=1\textwidth]{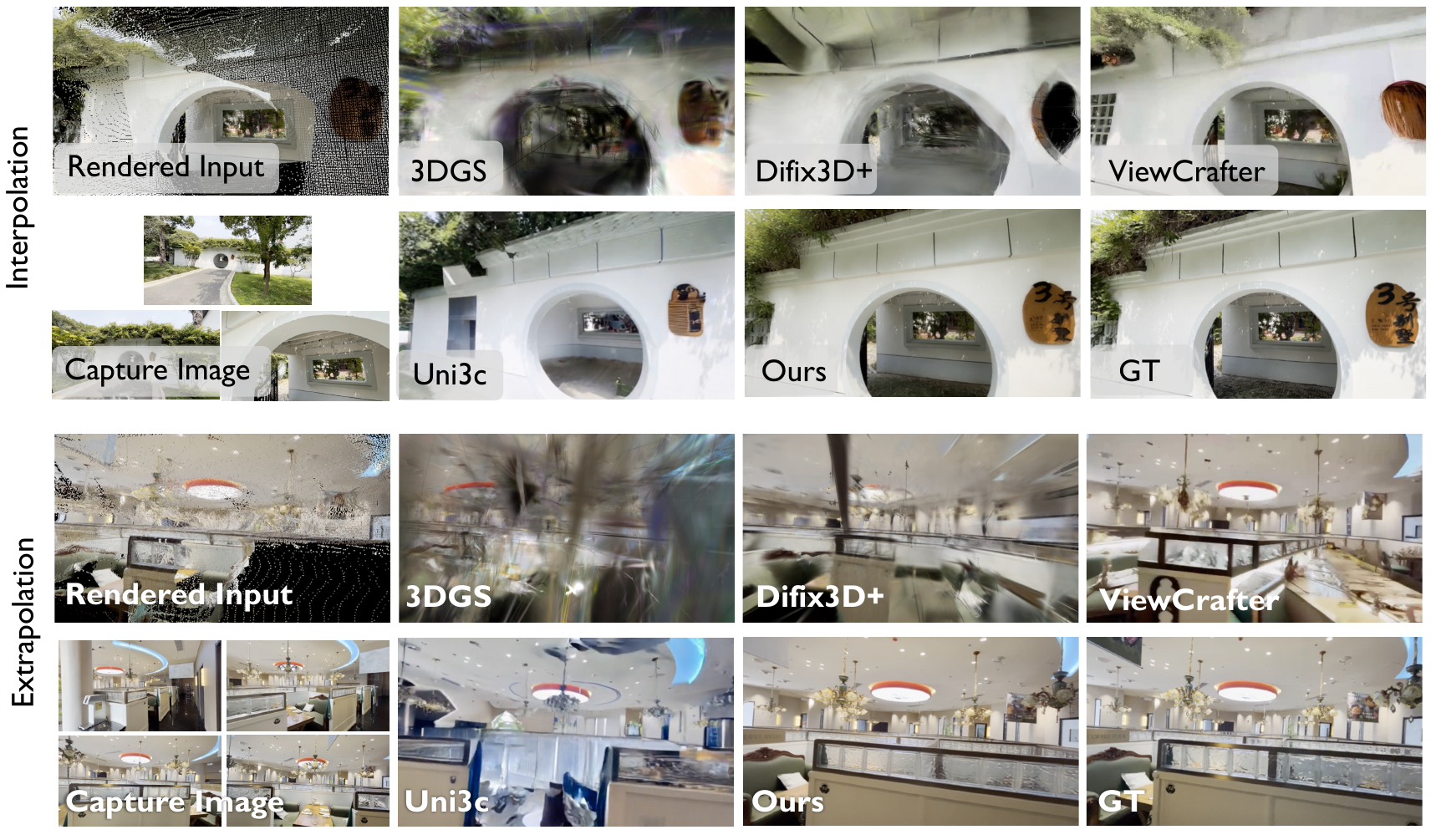}
   \caption{Quality Results on DL3DV Dataset~\cite{ling2024dl3dv}.}
   \label{fig:dl3dv}
   \vspace{-5mm}
\end{figure}

\subsection{Datasets} 
We train AnyRecon on the {DL3DV-10K}~\cite{ling2024dl3dv} dataset, a large-scale collection of high-quality 3D indoor and outdoor scenes. 
The original video sequences are partitioned into clips of 40 frames each at the resolution of $512\times 896$. 
To emulate diverse and irregular input scenarios while strengthening the model's generative priors, we employ a randomized conditioning sampling strategy. 
Specifically, for each clip, we fix the first frame as a base reference and randomly select $N \in [2, 4]$ additional conditioning views. 
To balance the model's ability to handle both narrow-baseline interpolation and wide-baseline synthesis, we sample these additional indices from either the first 20 frames (50\% probability) or the entire 40-frame window (50\% probability). The selected conditioning views $\mathcal{I}_{sel}$ are then processed by our feed-forward reconstruction module $\pi^3$~\cite{wang2025pi} to establish the initial 3D geometry memory $\mathcal{M}_{geo}$. 
Finally, we project the point-cloud observations from $\mathcal{M}_{geo}$ onto the target novel viewpoints $V_{novel}$ to generate the corresponding $I_{render}$ and visibility masks $M_t$, forming the complete training pairs for our geometry-controlled generative model.

\subsection{Implementation Details} 
We implement AnyRecon by fine-tuning the {Wan2.1-I2V-14B}~\cite{wan2025} model using {LoRA}~\cite{hu2022lora} with a rank of $32$. 
The training procedure is executed in three distinct stages to ensure stable convergence and efficient high-resolution synthesis. 
First, we perform full self-attention fine-tuning for $100$k iterations, allowing the model to adapt its internal generative priors to our geometry-controlled input space. 
Second, we transition to the sparse attention mechanism (\S\ref{sec:sparse}) and conduct a $10$k-iteration warm-up phase. 
Specifically, we configure the block sparse attention with a $2 \times 8 \times 8$ block size, restricting each frame to attend only to the selectively retrieved subset of geometry-aligned reference views $\mathcal{I}_{sel}$, alongside its $8$ preceding and $8$ succeeding adjacent views. 
This formulation enables the model to maintain long-range spatial consistency within the truncated receptive fields. 
Finally, we apply DMD2 distillation~\cite{yin2024improved} for an additional $30$k iterations, effectively compressing the denoising process into a 4-step sampling trajectory while preserving high-fidelity structural details. 
All experiments are conducted on 64 NVIDIA A800 GPUs using the AdamW optimizer with a constant learning rate of $1\times 10^{-4}$ for the initial stages and $1\times 10^{-5}$ during distillation.

\subsection{Comparison Results} 

\textbf{Metrics and Baselines.} We employ three widely-recognized metrics to quantitatively evaluate the synthesized results: Peak Signal-to-Noise Ratio (PSNR) for pixel-level accuracy, Structural Similarity Index (SSIM) for structural integrity, and Learned Perceptual Image Patch Similarity (LPIPS) for high-level perceptual quality. For a comprehensive comparison, we benchmark AnyRecon against three state-of-the-art diffusion-based 3D reconstruction and novel view synthesis methods: Difix3D+~\cite{wu2025difix3d+}, which focuses on geometry-refined image synthesis; ViewCrafter~\cite{yu2024viewcrafter}, which utilizes video diffusion priors for view interpolation; and Uni3C~\cite{cao2025uni3c}, a unified framework for cross-domain 3D consistency. These baselines represent the current frontier in leveraging generative models for sparse-view scenarios, providing a rigorous reference for assessing our model's advancements in spatial reasoning and efficiency.

\begin{figure}[!t]
   \centering
   \includegraphics[width=1\textwidth]{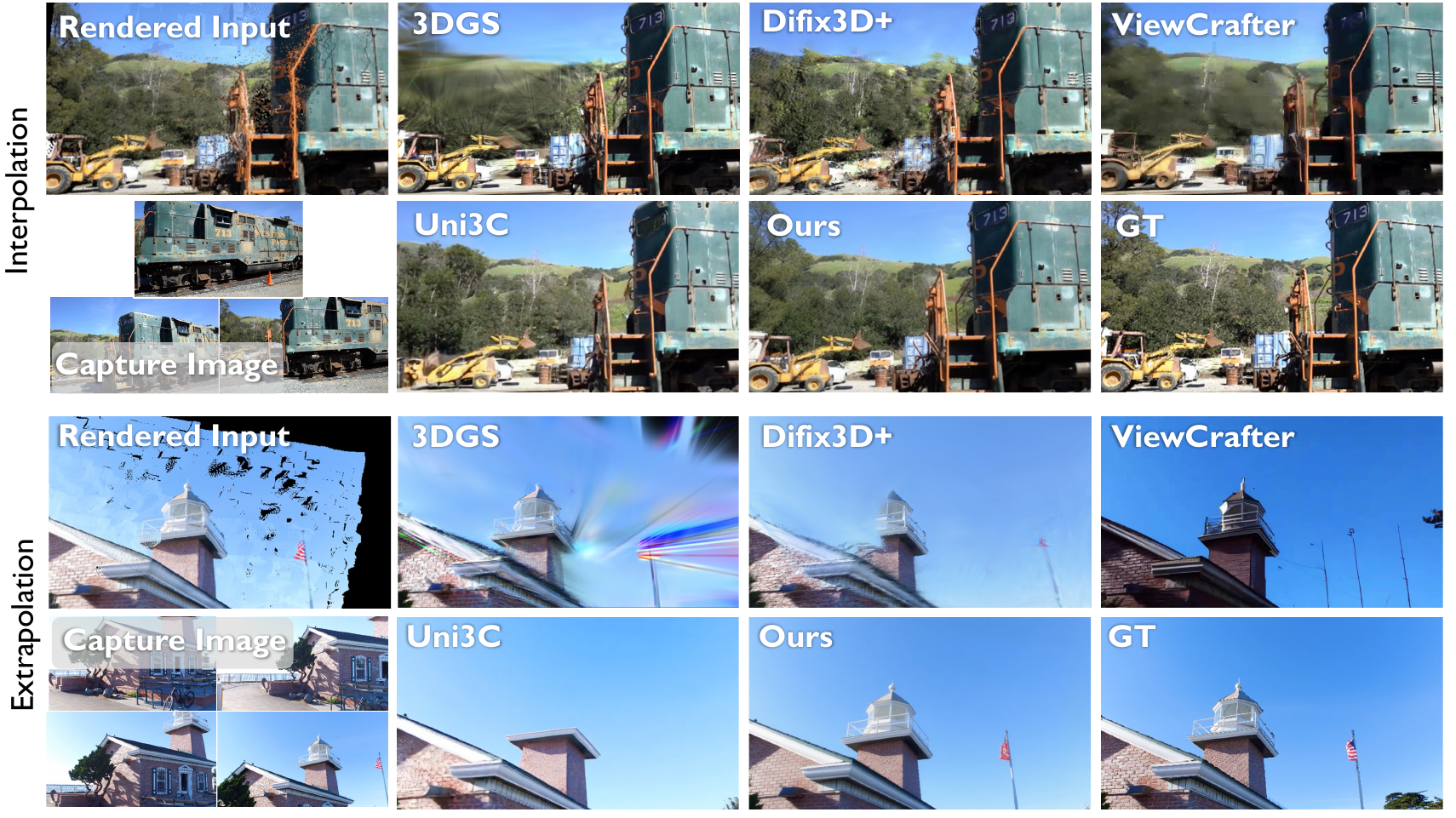}
   \caption{Quality Results on Tanks and Temples Dataset~\cite{Knapitsch2017}.}
   \label{fig:tank}
   \vspace{-5mm}
\end{figure}

\textbf{Evaluation Benchmarks.} To evaluate the generalization and robustness of AnyRecon, we conduct extensive testing on 10 scenes from {DL3DV-Evaluation} set~\cite{ling2024dl3dv}, and 5 scenes from {Tanks and Temples} Dataset~\cite{Knapitsch2017}. For each test sequence, we we sample 40 frames at a resolution of $512\times 896$ frames; specifically for the high-density Tanks and Temples sequences, we perform a $1/5$ temporal sub-sampling to ensure a challenging baseline for sparse-view reconstruction. Our evaluation is categorized into two distinct configurations: \textit{Interpolation} and \textit{Extrapolation}. In the Interpolation setting, we provide the $1^{st}$, $21^{st}$, and $40^{st}$ frames as captured views $\mathcal{V}$ to assess the model's sparse-view completion capability across large baseline gaps. In the Extrapolation setting, we provide the $1^{st}$, $11^{th}$, $21^{st}$, and $31^{st}$ frames as conditioning inputs to specifically test the model's generative synthesis ability in hallucinating visually and structurally coherent content for the unobserved tail of the trajectory.

\begin{table*}[t]
\centering
\small
\setlength{\tabcolsep}{4pt}
\caption{Quantitative comparison under interpolation and extrapolation settings.}
\resizebox{1\linewidth}{!}{
\begin{tabular}{l|ccc|ccc|c} 
\toprule
& \multicolumn{3}{c|}{Interpolation}
& \multicolumn{3}{c|}{Extrapolation} 
& \\ 
Method
& PSNR $\uparrow$ & SSIM $\uparrow$ & LPIPS $\downarrow$
& PSNR $\uparrow$ & SSIM $\uparrow$ & LPIPS $\downarrow$ 
& Time (s)$^*$ $\downarrow$ \\ 
\midrule
\rowcolor[HTML]{F2F2F2} \multicolumn{8}{l}{\textit{DL3DV}} \\ 
Difix3D+~\cite{wu2025difix3d+}      & 17.88 & 0.551 & 0.290 & 18.74 & 0.576 & 0.261 & 1200 \\
ViewCrafter~\cite{yu2024viewcrafter} & 15.86 & 0.463 & 0.394 & 15.51 & 0.459 & 0.406 & 170 \\
Uni3C~\cite{cao2025uni3c}            & 16.33 & 0.471 & 0.319 & 15.69 & 0.457 & 0.344 & 340 \\
\textbf{Ours}                        & \textbf{20.95} & \textbf{0.656} & \textbf{0.151} & \textbf{21.16} & \textbf{0.660} & \textbf{0.158} & \textbf{105} \\
\midrule
\rowcolor[HTML]{F2F2F2} \multicolumn{8}{l}{\textit{Tanks and Temples}} \\ 
Difix3D+~\cite{wu2025difix3d+}      & 19.43 & 0.629 & 0.163 & 18.67 & 0.594 & 0.190 & 1200 \\ 
ViewCrafter~\cite{yu2024viewcrafter} & 15.85 & 0.474 & 0.364 & 15.83 & 0.481 & 0.361 & 170 \\ 
Uni3C~\cite{cao2025uni3c}            & 16.77 & 0.514 & 0.263 & 16.54 & 0.502 & 0.274 & 340 \\ 
\textbf{Ours}                        & \textbf{20.37} & \textbf{0.639} & \textbf{0.158} & \textbf{20.30} & \textbf{0.629} & \textbf{0.181} & \textbf{105} \\ 
\bottomrule
\end{tabular}}
\begin{flushleft}
\footnotesize $^*$ All reported times represent the average inference duration for generating a 40-frame scene at a resolution of $512 \times 896$.The inference time for Difix3D+ is calculated based on its default 30-iteration refinement process (${\sim}40$s per iteration), excluding the initial 3DGS optimization time. The latency for all other methods, including ours, represents the duration of a single-pass diffusion model inference.
\end{flushleft}
\vspace{-0.5cm}
\label{tab:quantitative_interp_extra}
\end{table*}

Quantitative comparisons are summarized in Table~\ref{tab:quantitative_interp_extra}, and qualitative visualizations across different datasets are presented in Fig.~\ref{fig:dl3dv} and Fig.~\ref{fig:tank}. As illustrated, Difix3D+ fails to handle scenarios with large viewpoint gaps, often leaving significant artifacts inherited from sparse-view 3DGS reconstructions. While ViewCrafter and Uni3C leverage video diffusion priors, their inability to incorporate multiple conditioning frames during the diffusion process leads to generated views that do not strictly align with the captured observations. This results in cross-view geometric inconsistencies, loss of fine-grained details, and noticeable color shifts. In contrast, AnyRecon effectively leverages its global scene memory to complete missing regions in novel views based on the captured views, while simultaneously hallucinating plausible new content that maintains both structural integrity and appearance consistency. Moreover, AnyRecon achieves the best efficiency among all compared methods, requiring only 105 seconds per sequence. These results demonstrate that AnyRecon not only improves reconstruction fidelity but also significantly reduces inference latency, making it more practical for real-world applications.

\begin{table}[t]
\centering
\small
\setlength{\tabcolsep}{4pt} 
\caption{\textbf{Ablation study on temporal compression (TC) and inference efficiency.} We evaluate our model across various diffusion steps and attention strategies on the DL3DV Dataset~\cite{ling2024dl3dv} interpolation configuration.}
\label{tab:time_compression}
\resizebox{0.8\textwidth}{!}{
\begin{tabular}{l|ccc|cc}
\toprule
& \multicolumn{3}{c|}{\textbf{50 Diffusion Steps}} & \multicolumn{2}{c}{\textbf{4 Distilled Steps (w/o TC)}} \\
\cmidrule(lr){2-4} \cmidrule(lr){5-6}
Metric & Full TC & Paritail TC & w/o TC & Full Atten. & Sparse Atten. \\
\midrule
PSNR $\uparrow$  & 20.16 & 21.10 & 21.57 & 21.32 & 20.95 \\
SSIM $\uparrow$  & 0.616 & 0.661 & 0.687 & 0.673 & 0.656 \\
LPIPS $\downarrow$ & 0.179 & 0.153 & 0.140 & 0.148 & 0.151 \\
\midrule
Time (s)$^*$ $\downarrow$ & 210+(15) & 270+(15) & 1820+(15) & 140+(15) & {90}+(15) \\
\bottomrule
\end{tabular}}
\begin{flushleft}
\footnotesize {$^*$The reported Time (s) is formatted as ``DiT inference time + (15)'', where the 15s accounts for the encoder and decoder overhead. All values represent the average duration for generating a 40-frame video at a resolution of $512 \times 896$. Since the Full TC  requires input sequences of length $4n+1$, its runtime is measured on a 41-frame sequence and scaled by a factor of $40/41$ to align with the 40-frame baseline.} 
\end{flushleft}
\vspace{-0.5cm}

\end{table}

\subsection{Ablation Study}
\label{sec:ablation}
\textbf{Temporal Compression. } To verify the impact of different temporal processing strategies on the reconstruction quality, we conduct an ablation study with three configurations: full temporal compression, partial temporal compression (where only rendered maps are compressed while captured views remain uncompressed), and our proposed dense attention without temporal compression. The quantitative results and visual comparisons are presented in Table~\ref{tab:time_compression} and Fig.~\ref{fig:ablation tc}(c)(d)(e), respectively.

As illustrated in the results, both full and partial temporal compression lead to a noticeable degradation in visual fidelity. Specifically, these compression-based models struggle to preserve fine-grained structural details, such as the intricate metal grids shown in Fig.~\ref{fig:ablation tc}(c)(d), where the structures appear fractured or blurred. This is primarily because temporal down-sampling discards high-frequency spatial information that is crucial for thin-structure reconstruction. In contrast, our configuration without temporal compression effectively maintains the complete geometric details and sharp textures by attending to the original resolution of both rendered and captured views. This validates the necessity of preserving full temporal resolution to ensure high-fidelity scene synthesis in complex environments.

\textbf{Distillation and Sparse Attention.} We further evaluate the impact of our acceleration strategies, including model distillation and sparse Attention, as detailed in Table~\ref{tab:time_compression} and Fig.~\ref{fig:ablation tc}(e)(i)(j) . While the dense attention baseline (50 steps) achieves the highest reconstruction quality, its inference time ($1820$s) is prohibitively expensive for practical applications. By applying 4-step distillation, we significantly reduce the latency to $140$s with only a marginal drop in PSNR ($0.24$ dB). Furthermore, the integration of sparse Attention provides a substantial boost in efficiency, further compressing the inference time to $90$s—a $20\times$ speedup compared to the original dense baseline. Although the sparse constraints lead to a slight decrease in metrics (e.g., PSNR of $20.95$), the visual quality remains highly competitive, and the significant reduction in computational overhead makes AnyRecon much more viable for real-time 3D reconstruction tasks. This trade-off demonstrates that the combination of geometry-guided sparse attention and step distillation effectively balances high-fidelity synthesis with rapid deployment.

\begin{figure}[]
   \centering
   \vspace{-0.5cm}
   \includegraphics[width=1\textwidth]{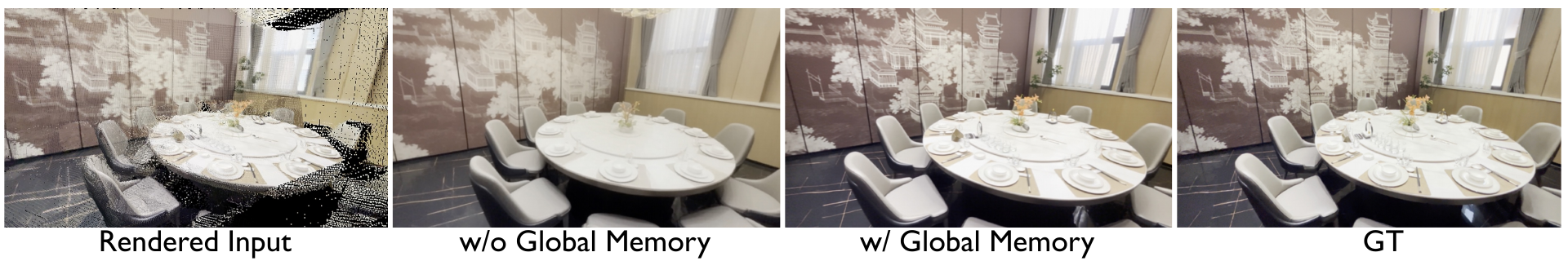}
   \caption{Quality comparison on global scene memory. }
   \vspace{-0.7cm}
   \label{fig:ablation_global}

\end{figure}

\textbf{Global Scene Memory.} 
To validate the necessity of the global scene memory, we conduct an ablation study on the DL3DV Dataset under the interpolation configuration. We compare our full model, which prepends three retrieved reference views into the global memory cache, against a baseline that only conditions the video diffusion model on a single initial frame. Note that to ensure a fair comparison, the explicit point-cloud guidance ($I_{render}$) in the baseline is still rendered using the geometry accumulated from all three views. Quantitative and qualitative comparisons are provided in Table~\ref{tab:ablation_global} and Fig.~\ref{fig:ablation_global}, respectively. Visually, relying solely on rendered point-cloud maps proves insufficient, as these intermediate renderings naturally suffer from projection artifacts—specifically, floating points, blurry boundaries, and inconsistent colors. Consequently, the baseline model struggles to recover high-fidelity textures, resulting in missing details on the tableware and noticeable color shifts on the background wall.
In contrast, by maintaining the raw captured views in the global scene memory, our full model allows the diffusion network to flexibly query uncorrupted, high-frequency textural details. This mechanism effectively suppresses geometric artifacts and successfully restores complex structures like the tableware, demonstrating the critical role of the global memory in preserving visual fidelity.

\begin{table}[h]

\centering
\caption{Quantitative comparison on global scene memory.}
\label{tab:ablation_global}
\setlength{\tabcolsep}{8pt}
\begin{tabular}{lccc}
\toprule
\textbf{Method} & \textbf{PSNR}$\uparrow$ & \textbf{SSIM}$\uparrow$ & \textbf{LPIPS}$\downarrow$ \\
\midrule
w/o Global Scene Memory &  20.18 & 0.634 & 0.205 \\
w/ Global Scene Memory & \textbf{20.95} & \textbf{0.656} & \textbf{0.151} \\
\bottomrule
\end{tabular}
\end{table}

\section{Limitation}
AnyRecon's performance depends on the quality of its 3D geometric memory. While resilient to minor inaccuracies—such as pose misalignments, noise, or artifacts—the framework requires basic structural coherence. In extreme cases with minimal view overlap, the initial reconstruction may fail, providing insufficient guidance for diffusion and resulting in suboptimal frame synthesis. 

\section{Conclusion}
We presented AnyRecon, a scalable and flexible framework designed for high-quality 3D reconstruction from sparse and irregular inputs. Addressing the limitations of existing diffusion-based methods in handling arbitrary views and large-scale scenes, we developed a novel video diffusion architecture that integrates explicit geometric control via point cloud renderings. By removing temporal compression and introducing a global memory cache, our model effectively maintains frame-level correspondence and supports unordered input conditioning. Furthermore, we proposed a geometry-aware conditioning strategy that establishes a closed loop between generation and reconstruction. Through the implementation of a 3D Geometry Memory and a geometry-driven view selection mechanism, AnyRecon enables robust, segment-by-segment reconstruction of complex, large-scale environments. Extensive experiments demonstrate that our approach significantly outperforms state-of-the-art baselines in view interpolation, extrapolation, and large-scene consistency, offering a practical solution for converting casual, sparse real-world captures into explorable 3D assets.
\label{sec:Conclusion}
%
%
\bibliographystyle{splncs04}
\bibliography{main}
\end{document}